\documentclass[journal,twoside,web]{ieeecolor}
\usepackage{jsen}
\usepackage{cite}
\usepackage{amsmath,amssymb,amsfonts}
\usepackage{algorithmic}
\usepackage{subfigure}
\usepackage{graphicx}
\usepackage{textcomp}
\usepackage{wrapfig}
\usepackage{soul}
\soulregister\ref7 
\soulregister\cite7
\newcommand{\tabincell}[2]{\begin{tabular}{@{}#1@{}}#2\end{tabular}}
\def\degree{${}^{\circ}$}
\def\BibTeX{{\rm B\kern-.05em{\sc i\kern-.025em b}\kern-.08em
    T\kern-.1667em\lower.7ex\hbox{E}\kern-.125emX}}
\definecolor{abstractbg}{rgb}{0.89804,0.94510,0.83137}
\begin{document}
\title{SGTBN: Generating Dense Depth Maps from Single-Line LiDAR}
\author{Hengjie Lu, Shugong Xu, \IEEEmembership{Fellow, IEEE}, and Shan Cao, \IEEEmembership{Member, IEEE}
}

\IEEEtitleabstractindextext{%
\fcolorbox{abstractbg}{abstractbg}{%
\begin{minipage}{\textwidth}%
\begin{wrapfigure}[12]{r}{3in}%
\includegraphics[width=2.95in]{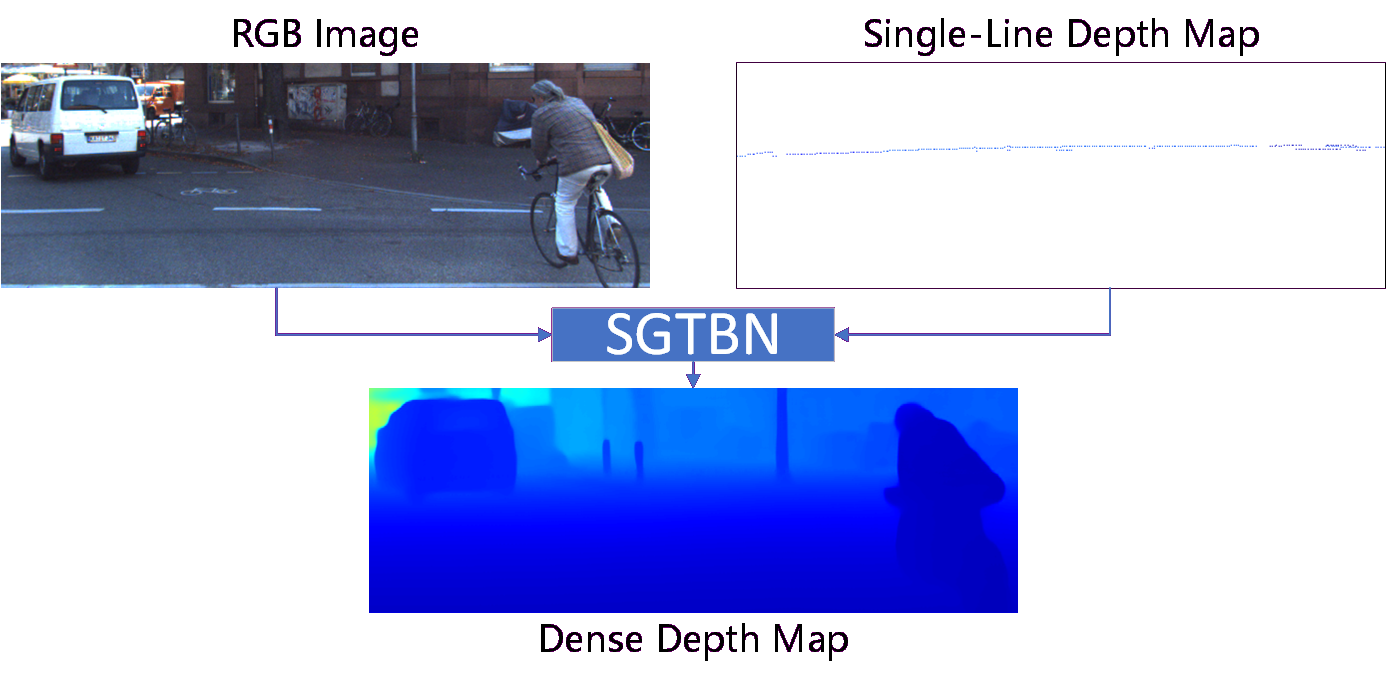}%
\end{wrapfigure}%
\begin{abstract}
Depth completion aims to generate a dense depth map from the sparse depth map and aligned RGB image. However, current depth completion methods use extremely expensive 64-line LiDAR(about \$100,000) to obtain sparse depth maps, which will limit their application scenarios. Compared with the 64-line LiDAR, the single-line LiDAR is much less expensive and much more robust. Therefore, we propose a method to tackle the problem of single-line depth completion, in which we aim to generate a dense depth map from the single-line LiDAR info and the aligned RGB image. A single-line depth completion dataset is proposed based on the existing 64-line depth completion dataset(KITTI). A network called Semantic Guided Two-Branch Network(SGTBN) which contains global and local branches to extract and fuse global and local info is proposed for this task. A Semantic guided depth upsampling module is used in our network to make full use of the semantic info in RGB images. Except for the usual MSE loss, we add the virtual normal loss to increase the constraint of high-order 3D geometry in our network. Our network outperforms the state-of-the-art in the single-line depth completion task. Besides, compared with the monocular depth estimation, our method also has significant advantages in precision and model size.
\end{abstract}

\end{minipage}}}

\maketitle

\section{Introduction}
\label{sec:introduction}
\IEEEPARstart{I}{n} recent years, accurate and dense depth maps have become increasingly important for various tasks, such as augmented reality, unmanned aerial vehicle control, robot control, and autonomous driving. Due to the strong interference and far range in the outdoors, devices such as near infrared cameras and structured light cameras will be ineffective. With the far measurement distance and strong anti-interference capability, LiDAR has become the mainstream solution for outdoor depth measurement. However, the limited scan line of LiDAR also leads to the high sparsity of its depth map. For instance, for the 64-line LiDAR Velodyne HDL-64E, which is the LiDAR sensor with 64 scan lines, its depth map only contains around 4$\%$ valid pixels\cite{b1}. Such a sparse depth map is far from enough for practical applications.

At present, the common solution to this problem is using the sparse depth map captured by the 64-line LiDAR and the aligned RGB image to generate the dense depth map. This method is commonly referred to as depth completion. With this method, we can obtain dense depth maps sufficient for practical applications. However, the 64-line LiDAR is very expensive. For instance, the 64-line LiDAR Velodyne HDL-64E costs about \$100,000. Such high costs will limit its use in industry. The single-line LiDAR, which is the LiDAR sensor with one scan line, costs much less compared with the 64-line LiDAR. For instance, the single-line LiDAR P+F R2000 costs only about \$2,000. As a result, we consider using single-line LiDAR instead of 64-line LiDAR in the depth completion to save cost.

Our goal is to use the single-line depth map and the aligned RGB image to generate the dense depth map. Through this way, accurate and dense depth maps can be obtained at a lower cost. At present, the most commonly used outdoor dataset for depth completion is KITTI \cite{b2} dataset. However, the KITTI is used in the 64-line depth completion task and there is no dataset like KITTI used in the single-line depth completion task. Therefore, we decided to generate a single-line depth completion dataset from KITTI by converting 64-line depth maps to single-line depth maps.

\begin{figure}[htb]
  \centering
  \subfigure[64-line depth map]{
  \includegraphics[width=0.95\linewidth]{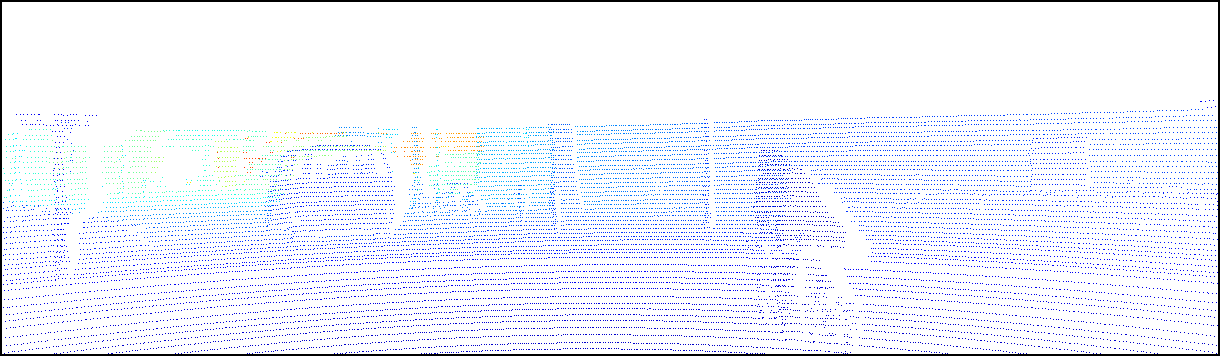}
  }
  \subfigure[single-line depth map]{
  \includegraphics[width=0.95\linewidth]{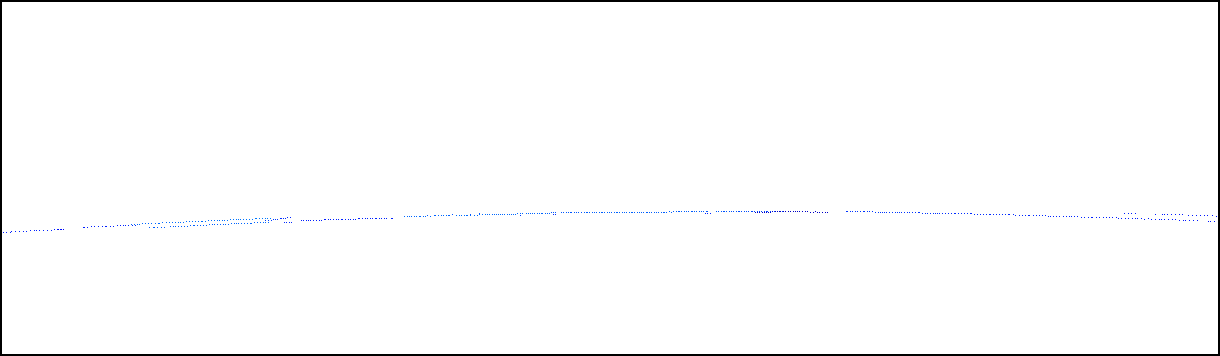}
  }
  \caption{The qualitative comparison between the (a) 64-line depth map and the (b) single-line depth map. The single-line depth map is generated by processing the 64-line depth map. Moreover, we converted depth maps from 16-bit grayscale to RGB for better observation.}
  \label{fig:sparse depth maps comparison}
\end{figure}

As illustrated in Figure~\ref{fig:sparse depth maps comparison}, we can find that the single-line depth map is much sparser than the 64-line depth map. Therefore, the single-line depth completion is much more difficult than 64-line depth completion and relies more on semantic info in the RGB images.

To accomplish this task, we propose a novel network called Semantic Guided Two-Branch Network(SGTBN). Firstly, we use the combination of the global and local branches to extract and fuse global and local info. Secondly, the semantic guided depth upsampling module is proposed to make better use of the semantic info of the RGB image. Thirdly, except for the usual MSE loss, we add the virtual normal loss. With the help of this extra supervision info, the network will pay more attention to the scene's 3D geometric structure.

We use the dataset generated from KITTI to evaluate our network's performance in the single-line depth completion task. Our network is compared with advanced networks designed for depth completion. Although the networks they designed were previously used for 64-line depth completion, they can also be used for single-line depth completion with some modification, which can make a fair comparison. Experimental results on the dataset we generate demonstrate that our network achieves state-of-the-art performance in the single-line depth completion task. Specifically, our method can achieve the relative error of 4.68$\%$ without pre-training. We also study the multi-modal data fusion form in the single-line depth completion task. Besides, we also compare our method with the monocular depth estimation. The result shows that compared with the state-of-the-art monocular depth estimation method, our method has significant advantages in precision and model size. 

In summary, the contributions of this paper are:
\begin{enumerate}
\item We propose a method to generate dense depth maps from the single-line LiDAR and RGB camera. In our method, the single-line depth completion dataset can be obtained by processing the KITTI depth completion dataset(64-line). The relative error of our method is only 4.68$\%$, which means that for the depth of $1m$, the error of our method is only $\pm 4.68cm$, which is of great practical value.
\item We design a novel network called Semantic Guided Two-Branch Network(SGTBN) for single-line depth completion. 
\item Through experiments, we find that our network achieves state-of-the-art performance in the single-line depth completion task. Besides, compared with the monocular depth estimation, our method also has significant advantages in precision and model size. We also designed several ablation experiments to verify the rationality of our network design.
\end{enumerate}

\section{Related Work}
At present, the common generation methods of dense depth maps can be divided into two categories: depth completion and monocular depth estimation. In the previous work, depth completion used the RGB image and aligned 64-line depth map to generate the dense depth map. Monocular depth estimation only uses the single RGB image as input to estimate the dense depth map.

\textbf{Depth Completion. }The previous researches on depth completion were based on the 64-line depth map. So the related depth completion work discussed here is also based on the 64-line depth map. Traditional methods often use hand-crafted features to complete invalid values of depth maps \cite{b3, b4, b5}.  Recently, the methods based on deep learning have shown promising performance on depth completion. Uhrig \emph{et al.} \cite{b6} proposed the sparsity-invariant convolution to make the network invariant to the level of sparsity. Qiu \emph{et al.} \cite{b7} estimates surface normals to assist in generating the final dense depth map. Eldesokey \emph{et al.} \cite{b8} propose strategies that can determine the conﬁdence from the convolution operation and propagate it to consecutive layers. Yan \emph{et al.} \cite{b9} propose strategies to process, downscale, and fuse sparse features. Van Gansbeke \emph{et al.} \cite{b10} propose a framework to extract both local and global info and fuse them with confidence maps.

\textbf{Monocular Depth Estimation. }Traditional methods often use hand-crafted features extracted from the monocular images to predict the depth maps\cite{b11}. Eigen \emph{et al.} \cite{b12} is the first team that uses deep learning to accomplish monocular depth estimation. They present a two-stage network to address this problem. Yin \emph{et al.} \cite{b13} designed a virtual normal loss term to add additional 3D geometric constraints. Lee \emph{et al.} \cite{b14} propose the local planar guidance layers to make full use of the encoded feature.

Although great progress has been made in 64-line depth completion and monocular depth estimation, these two methods still have great disadvantages. The 64-line depth completion requires the extremely expensive 64-line LiDAR which significantly restricts its application in the industry. Although the cost of monocular depth estimation is low, its accuracy is too low to be applied in practice and its model size is usually huge. Therefore, we propose a method that can be applied to single-line LiDAR to meet the practical application, with low cost for high precision.

\section{Method}

We propose a network called Semantic Guided Two-Branch Network(SGTBN) for the single-line depth completion. In this section, we will describe this network in detail. In our network, global and local branches are used to extract corresponding info to aid in generating dense depth maps more accurately. For the multi-modal data fusion method in the single-line depth completion, we decided to use the late fusion method. The semantic guided depth upsampling module(SGDUM) is used to make better use of the semantic info in RGB images. Besides, in addition to the MSE loss, we also use the virtual normal loss(VNL) to enforce the 3D geometric constraints.

\subsection{Overall Structure}

\begin{figure*}[htb]
  \centering
  \includegraphics[width=1\linewidth]{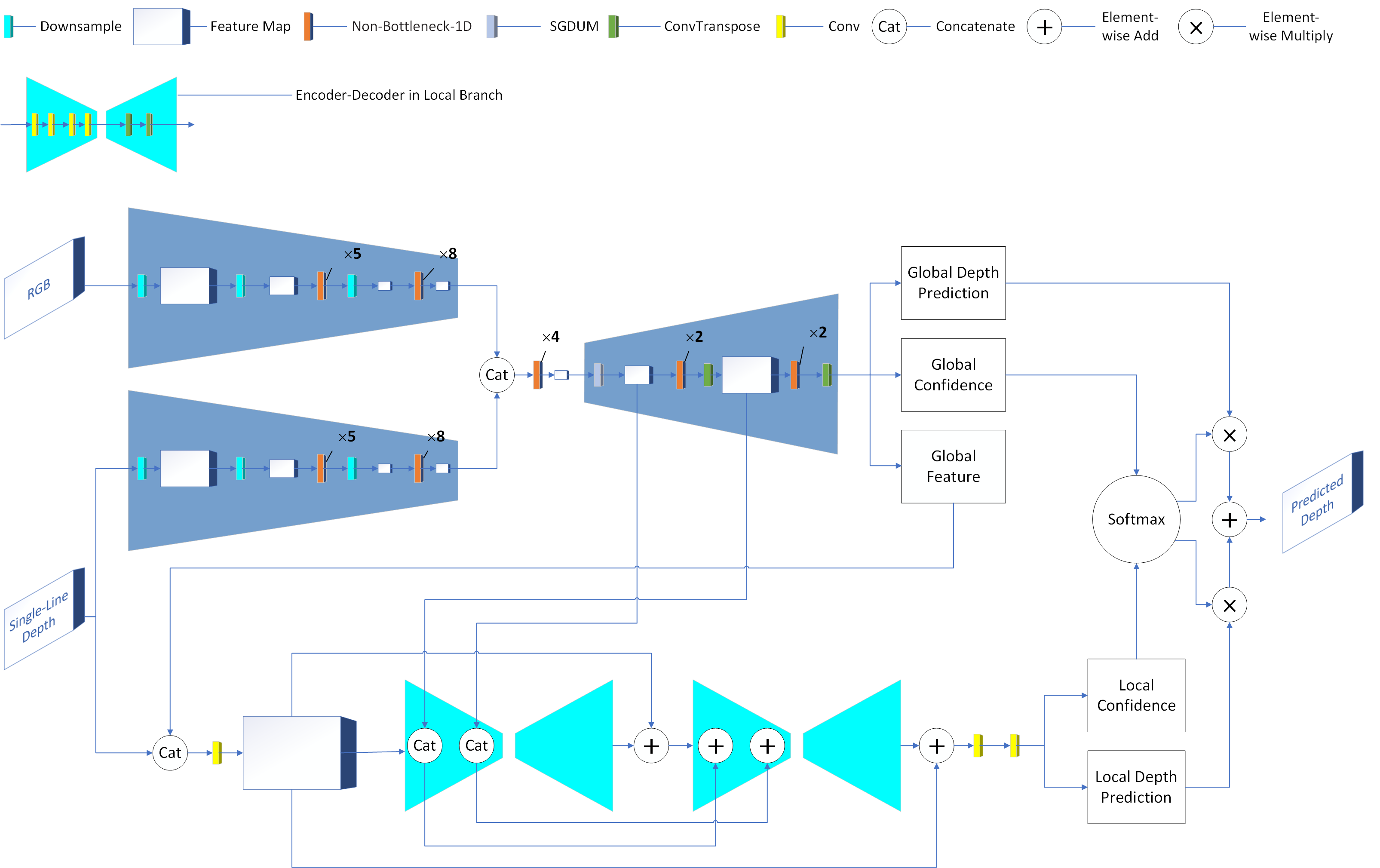}
  \caption{The illustration of the proposed network. There are two branches: the global branch(top) and the local branch(bottom). Through two branches, we can get two depth maps and the corresponding confidence maps. The confidence maps can become weight maps after the softmax function. The final depth map is the weighted sum of depth maps and their corresponding weight maps in two branches.}
  \label{fig:overall structure}
\end{figure*}

As shown in Figure~\ref{fig:overall structure}, our overall structure can be divided into two branches: global and local branches. The top branch is global and the bottom branch is local. The idea of global and local branches comes from \cite{b10}.

The global branch uses the RGB image and single-line depth map as input. The local branch uses the single-line depth map and global feature which the global branch extracts as input. With the semantic information directly provided by the RGB image, the global branch can achieve better results in areas far away from valid values in the depth map. The local branch focuses on the valid value areas in the depth map and predicts the depth better around those areas. Besides, the global branch can also achieve better results on the valid values, which is mainly due to two reasons. On one hand, the global branch can directly use the semantic information of RGB images to correct the inherent errors in the single-line depth map. On the other hand, the global branch has a separate encoder to process the single-line depth map, unlike the local branch which processes it after fusing the single-line depth map with the global features. In a word, the global branch relies on the global semantic info provided directly by RGB images to achieve better results in areas that far from valid values and on valid values. Local branches can achieve better results in areas close to valid values.

In the global branch, we use two-modal data as input: RGB image and single-line depth map. For them, our strategy is to use two encoders to extract the features of the RGB image and depth map respectively and then fuse them. This type of strategy represents the late fusion approach to process multi-modal data. Compared with the other early fusion strategy, that is, directly using one encoder to process the RGB-D image, our strategy can use different networks to process the different modal data and achieve better results. The Non-bottleneck-1D in the global branch is proposed in the ERFNet \cite{b15}. It is the variant of Non-bottleneck in ResNet \cite{b16}. It replaces the $3\times3$ convolution in Non-bottleneck with $3\times1$ and $1\times3$ convolution. Non-bottleneck-1D reduces parameters and gains greater non-linearity. The downsampling module in the encoder of the global branch combines convolution and pooling operations. This module concatenates the feature operated by the convolution(stride=2) and max pooling(stride=2). In the single-line depth completion task, the depth info in the sparse depth map is too little. Therefore, we replace part of the transposed convolution with the semantic guided depth upsampling module(SGDUM). With this module, we can generate the dense depth map more accurately by making better use of semantic info.

In the local branch, we use two sets of encoder-decoder. Since the local branch doesn't need to extract the semantic info of the RGB image, the encoder-decoders in it only use ordinary convolution and transpose convolution to compress and restore features. Although the local branch focuses on the valid value areas in the depth map, it also needs off-the-shelf semantic info to guide local depth completion. So we add the global feature to the input and add the feature from the global branch to the first encoder. What's more, several residual structures are used in the local branch.

Through the global branch and local branch, we can get five feature maps: global depth prediction, global confidence, global feature, local depth prediction, and local confidence. The global feature is used to guide local depth completion. Two confidence maps can be the weight maps of each branch through the softmax function. Finally, we fuse the prediction depth maps of the global and local branches and their corresponding weight maps to obtain the final predicted depth map. This process of fusing two prediction depth maps based on confidence is similar to \cite{b7}. The final predicted depth map $D_{Final}$ is defined as:
\begin{align}
D_{Final} = & \frac{e^{C_{Global}}}{e^{C_{Global}} + e^{C_{Local}}} \times D_{Global} \nonumber \\ & + \frac{e^{C_{Local}}}{e^{C_{Global}} + e^{C_{Local}}} \times D_{Local}
\end{align}
where $D$ represents Depth and $C$ represents Confidence.

\begin{figure*}[htb]
  \centering
  \includegraphics[width=1\linewidth]{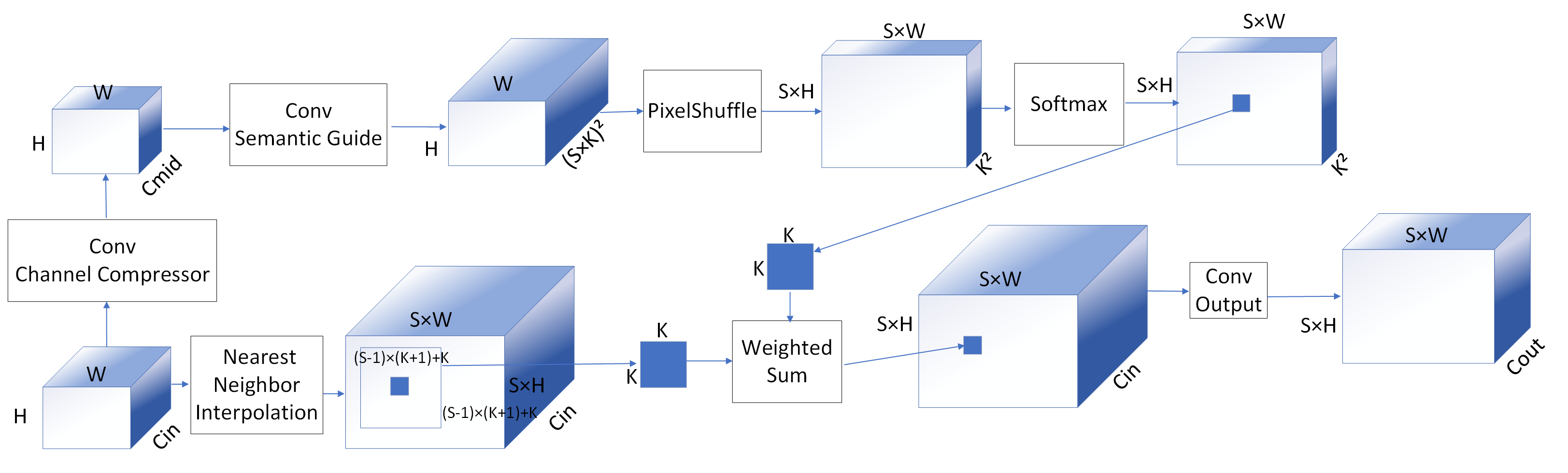}
  \caption{The structure of the SGDUM Module. This module can be divided into two branches. The top is used to generate semantic guided kernels. The bottom uses semantic guided kernels to realize upsampling. The (C, H, W) in this figure represents channel, height, and width respectively. The K(K=5) represents the kernel size. In this module, the feature map is upsampled by a factor of S(S=2).}
  \label{fig:SGDUM}
\end{figure*}

\subsection{Semantic Guided Depth Upsampling Module}

Compared with the previous 64-line depth completion, the depth info in the single-line depth completion task is more lacking. Because of this situation, we propose the Semantic Guided Depth Upsampling Module(SGDUM). This module is mainly inspired by CARAFE \cite{b17} which is confirmed effective in object detection, instance segmentation, semantic segmentation, and image inpainting. Different from the transposed convolution using the same kernel in the entire feature map, our module can use different kernels in different regions according to the semantic info of the feature map. Therefore, even when depth info is extremely scarce, our network can still generate accurate and dense depth maps by making better use of semantic info in the upsampling stage of the decoder.

As shown in Figure~\ref{fig:SGDUM}, the SGDUM can be divided into two branches. The top branch is used to generate semantic guided kernels. The bottom branch uses the semantic guided kernels to upsample the feature map.

In the top branch, generating semantic guided kernels can be divided into three stages. Firstly, the channel of feature will be compressed through a $1\times1$ convolution layer to reduce the parameters and computational cost in the following steps($C_{mid} = C_{in} \div 4$). Secondly, we use $3\times3$ convolution and pixelshuffle to generate the corresponding kernels based on the semantic info of each location of the feature map. Finally, we realize the channel-wise normalization through the softmax function. After this, we can get the feature map with $K^{2} \times (S \times H) \times (S \times W)$, where $K$ represents the size of semantic guided kernels. Each position in this feature map has $K \times K$ parameters which will be the fusion weights when upsampling.

In the bottom branch, we will use the semantic guided kernels generated by the top branch to implement the upsampling. This process can also be divided into three stages. Firstly, we use the nearest neighbor interpolation to upsample the feature map. After this stage, the resolution of the feature here is the same as the feature generated by the top branch. Secondly, we use the sliding window to gradually obtain the neighbors at different locations in the feature map. In this stage, the kernel size of the sliding window is $K \times K$ and the dilation is $S$. Through these operations, we can get the $K \times K$ neighbors in each position of each channel. Then we can use the corresponding $K \times K$ semantic guided kernel obtained by the top branch to fuse them. The same location in different channels will use the same semantic guided kernel. Finally, a $1 \times 1$ convolution is used to convert the number of channels.

\subsection{Loss}
Most depth completion methods only used MSE(Mean Square Error) as the loss function. These methods usually ignore the geometric info in the 3D space. So we add an extra Virtual Normal Loss(VNL) to compensate for this weakness. The virtual normal loss is proposed by Yin \emph{et al.} \cite{b13} in the monocular depth estimation. Virtual normal loss can make our network pay more attention to 3D geometric info in generating dense depth maps, thus greatly improving the accuracy in single-line depth completion.

To add virtual normal loss, we firstly randomly select several point groups in the 3D space of prediction and ground truth. Each group has 3 points which are restricted to be non-colinear to establish a virtual plane. Since the three points are randomly selected, the virtual plane may not contain any physical meaning. Secondly, we generate the virtual normal from the virtual plane. Finally, We compute the virtual normal loss between prediction and ground truth. The virtual normal loss is defined as:
\begin{align}
L_{VN} = \frac{1}{N}(\sum\limits_{i=1}^{N}{||{n_i}^{pred}-{n_i}^{gt}||}_1)
\end{align}
where the N is the number of groups and the $n_i$ is the virtual normal of each group.

The final loss we use includes two parts: MSE Loss and Virtual Normal Loss. The MSE Loss is used to enforce pixel-wise depth supervision and the Virtual Normal Loss is used to enforce 3D geometric supervision. The final loss is defined as:
\begin{align}
L_{Final} = L_{MSE} + \lambda L_{VN}
\end{align}
where $\lambda(\lambda = 100)$ is used to balance two losses.

Furthermore, to make the network predict the global and local depth better, we add the intermediate supervision to the global and local branches. The total loss is defined as:
\begin{align}
L_{Total} = & L_{Final}(\widehat{y}_{Out}, y) \nonumber \\ & + w_{1} \cdot L_{Final}(\widehat{y}_{Global}, y) \nonumber \\ & + w_{2} \cdot L_{Final}(\widehat{y}_{Local}, y)
\end{align}
where $\widehat{y}$ represents the prediction and $y$ represents the ground truth. The weights $w_{1}$ and $w_{2}$ are both equal to 0.1.

\section{Experiment}
We use a single-line depth completion dataset to validate the effectiveness of our network. The dataset will be described in detail in the following section. For fairness, all the models in our experiments are trained from scratch (without any pre-trained weights). In all experiments, we use RMSE as the primary metric because it is also the main metric on the KITTI depth completion benchmark(64-line depth completion).

\subsection{Dataset}
KITTI depth completion dataset \cite{b6} is a large real-world outdoor dataset that is used to address the 64-line depth completion. In the KITTI depth completion dataset, the 64-line depth maps that contain about 4$\%$ valid pixels \cite{b1} and aligned RGB images are used as input, and the semi-dense depth maps that contain about 30$\%$ valid pixels \cite{b9} are used as ground truth. The ground truth is semi-dense because the dense depth annotation is not available in the real world \cite{b18}.

We generate the single-line depth completion dataset by converting 64-line depth maps in KITTI to single-line depth maps. The method of generating single-line depth maps from 64-line depth maps comes from Yan \emph{et al.} \cite{b9}. They use this method to generate the depth maps with fewer scan lines such as 32-line, 16-line, and so on. Although they also generate the sparser depth maps from 64-line depth maps, their approach using sparser depth maps is fundamentally different from ours. Their goal is to verify the robustness of the model which is trained on KITTI(64-line). So they only use the depth maps with fewer lines during inference and use 64-line depth maps during training. Our goal is to use our model on single-line LiDAR. So we use single-line depth maps both in training and inference. Their approach has the domain gap between training and inference, so even if the model is robust, it still can not be applied to the LiDAR with fewer scan lines. The domain gap will reduce the performance of the model greatly. In the single-line depth completion, the error of our approach is about one-tenth of theirs on RMSE and one-fifteenth of theirs on MAE. That's because unlike theirs our approach doesn't have the domain gap between training and inference.

The method of generating single-line depth maps from 64-line depth maps can be divided into four steps. Firstly, we need to convert the 64-line depth map to the point cloud map. Secondly, we calculate the vertical angles for each 3D point. The vertical angles is defined as $\theta = arcsin (\frac{z}{\sqrt{x^2+y^2+z^2}})$. Thirdly, we quantify $\theta$ into 64 levels with the interval of 0.4\degree according to the SPEC from Velodyne to assign each point to the corresponding scan lines. Finally, we select the middle scan line to generate the single-line depth map.

Since the KITTI depth completion dataset is used for 64-line depth completion, we can not use the official website of KITTI benchmarks to evaluate our network. So we need to redivide the KITTI depth completion dataset to evaluate by ourselves. The KITTI depth completion dataset which provides ground truth consists of three sets: training set, validation set, selection validation set. There is no intersection between the training set and validation set. The selection validation set is the subset of the validation set. We separate the selection validation set from the validation set and use the selection validation set as the test set. Through this way, we can get the mutually disjoint training set, validation set, and test set.

\subsection{Implementation Details}

\begin{figure*}[htb]
  \centering
  \subfigure[RGB Images]{
  \includegraphics[width=0.45\linewidth]{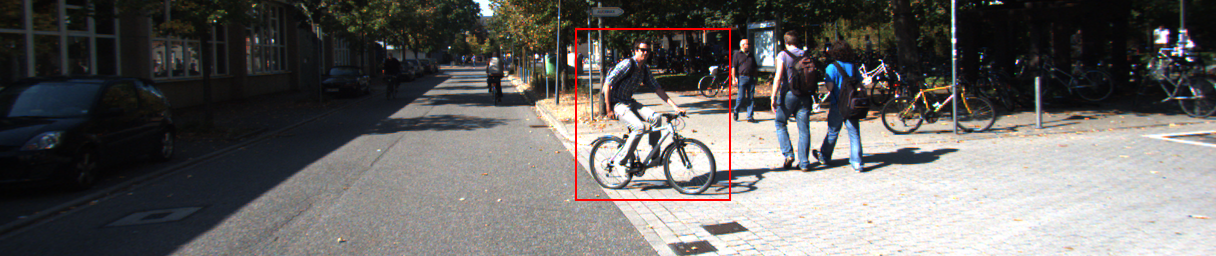}
  \includegraphics[width=0.45\linewidth]{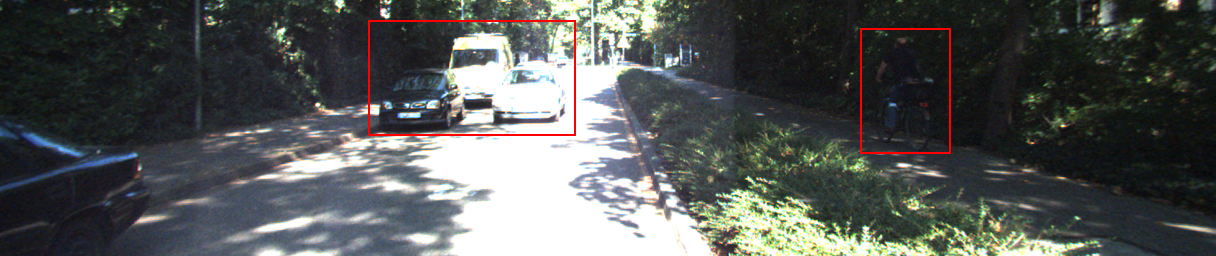}
  }
  \subfigure[Single-Line Depth Maps]{
  \includegraphics[width=0.45\linewidth]{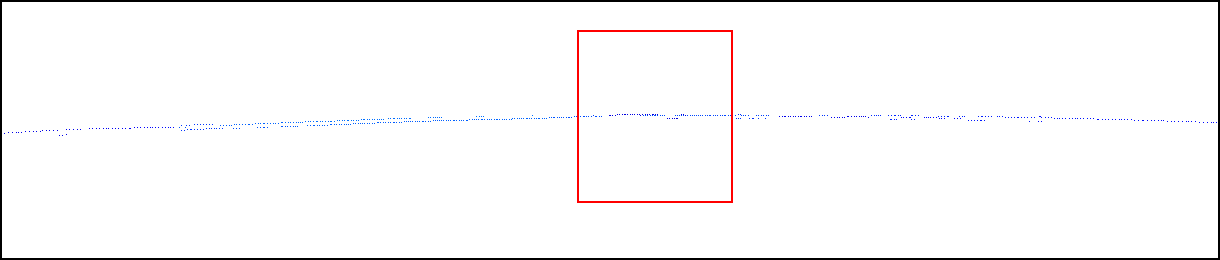}
  \includegraphics[width=0.45\linewidth]{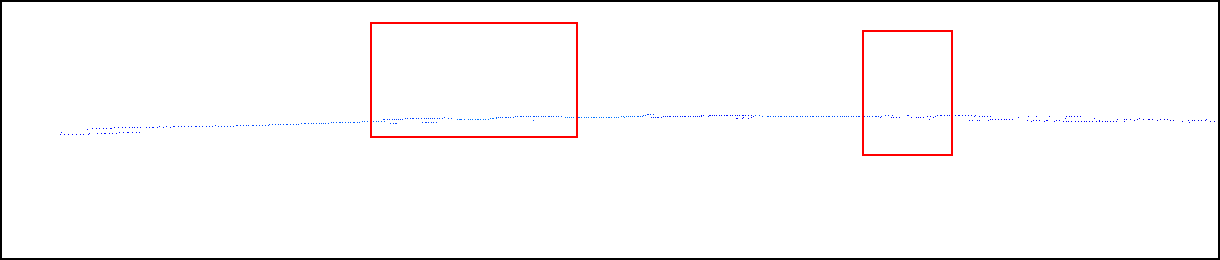}
  }
  \subfigure[Fusion Maps(RGB \& Depth)]{
  \includegraphics[width=0.45\linewidth]{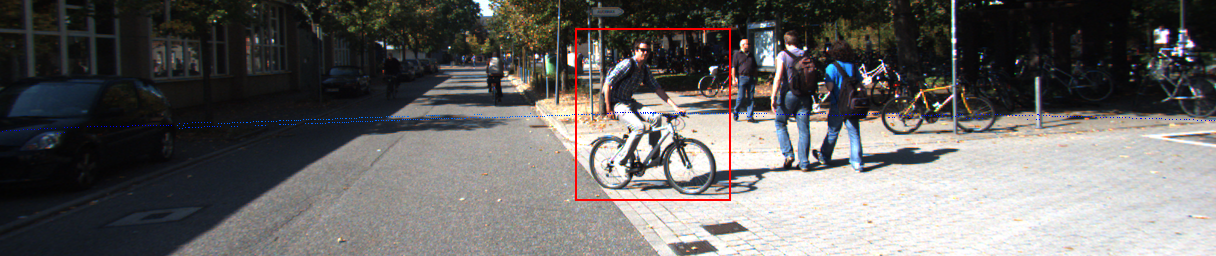}
  \includegraphics[width=0.45\linewidth]{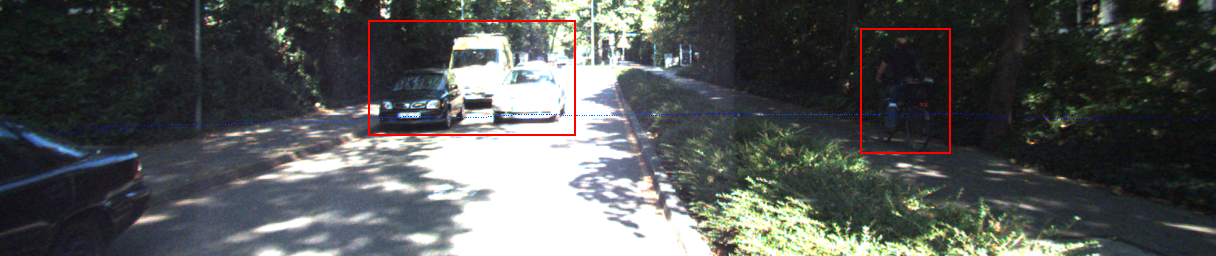}
  }
  \subfigure[Revisiting \cite{b9}]{
  \includegraphics[width=0.45\linewidth]{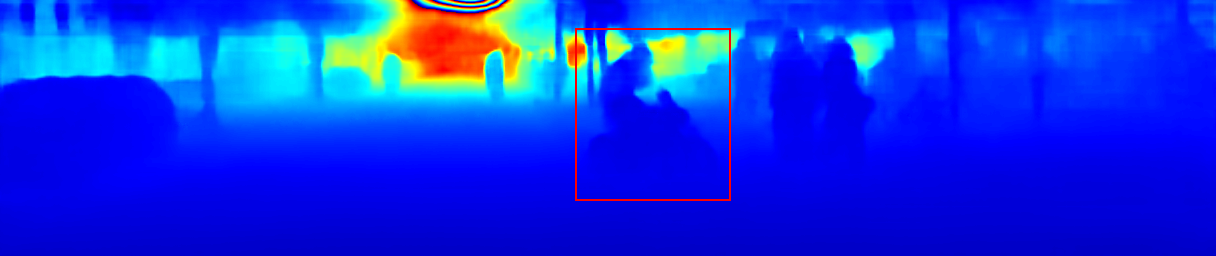}
  \includegraphics[width=0.45\linewidth]{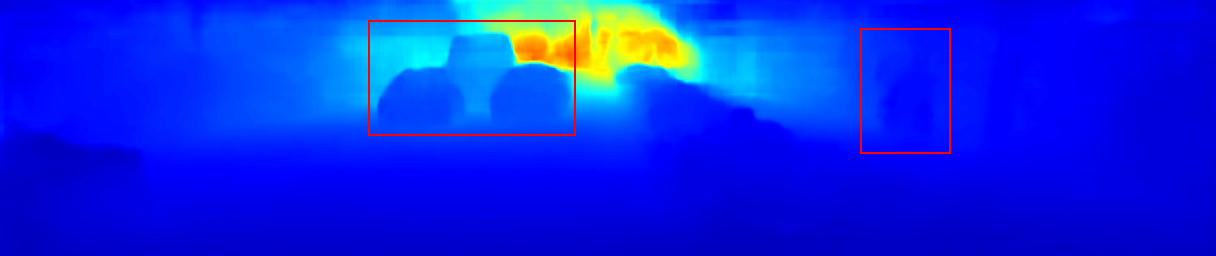}
  }
  \subfigure[RGB-guide$\&$certainty \cite{b10}]{
  \includegraphics[width=0.45\linewidth]{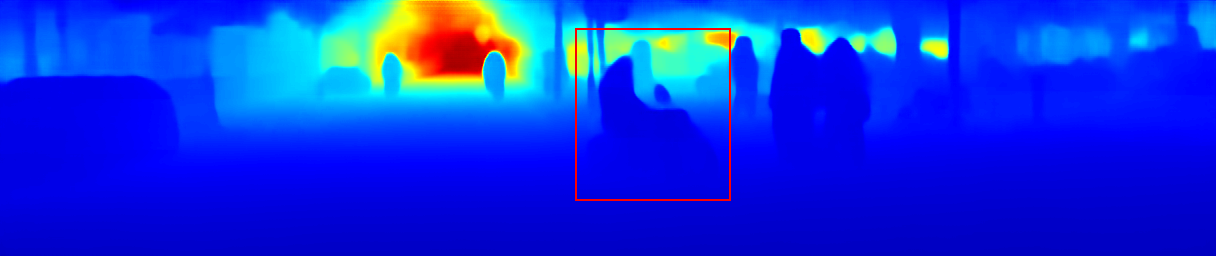}
  \includegraphics[width=0.45\linewidth]{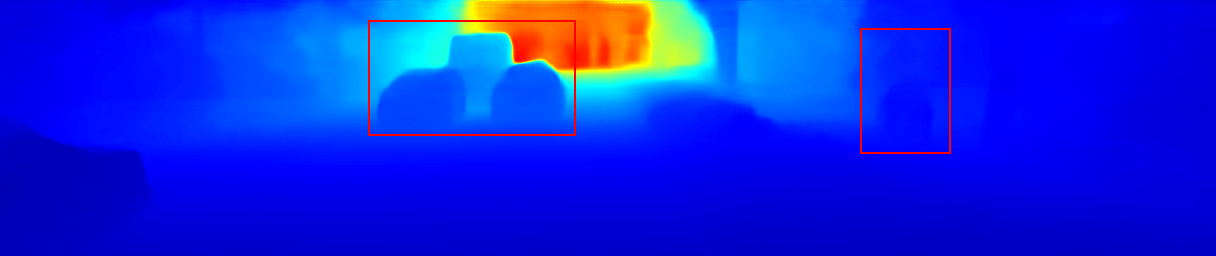}
  }
  \subfigure[Ours]{
  \includegraphics[width=0.45\linewidth]{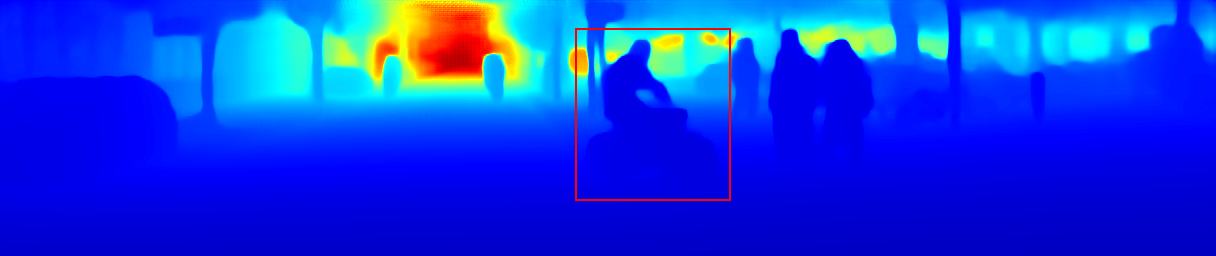}
  \includegraphics[width=0.45\linewidth]{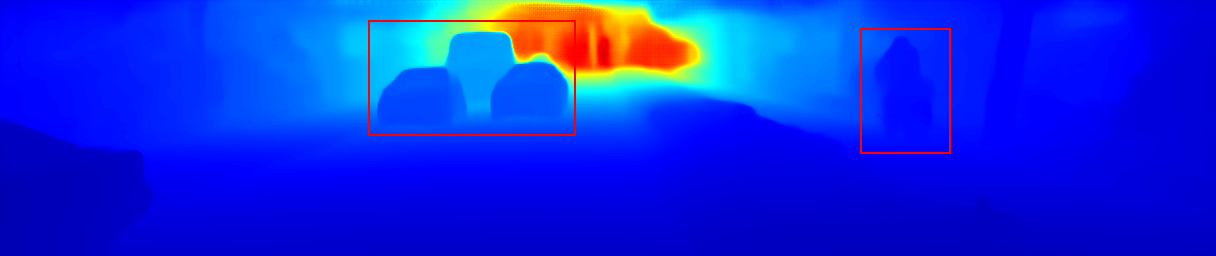}
  }
  \caption{Qualitative comparison with other methods on single-line depth completion. There are two sets of qualitative results here. The RGB image and single-line depth map are used as input in all methods here. The fusion maps represent the overlay of the aligned single-line depth maps on the RGB images to observe them better.}
  \label{fig:qualitative results on single-line depth completion}
\end{figure*}

\begin{table*}[htb]\normalsize
\caption{Quantitative comparison on single-line depth completion task}
\label{single-line depth completion comparison}
\begin{center}
\begin{tabular}{|l|c|c|c|c|c|c|}
\hline
\tabincell{l}{Network} & \tabincell{c}{RMSE \\ (mm) } & \tabincell{c}{MAE \\ (mm) } & \tabincell{c}{iRMSE \\ (1/km) } & \tabincell{c}{iMAE \\ (1/km) } & \tabincell{c}{sqErrorRel \\ (percent)} & \tabincell{c}{absErrorRel \\ (percent) } \\
\hline
NConv-CNN \cite{b8}  & 6607.35 & 3035.96 & 20.87 & 13.04 & 7.45 & 16.33 \\
Revisiting \cite{b9}  & 3356.27 & 1374.82 & 9.15 & 5.44 & 1.53 & 6.88 \\
RGB-guide$\&$certainty \cite{b10}  & 3205.47 & 1260.87 & 7.43 & 4.26 & 1.21 & 5.68 \\
Ours  & \textbf{2921.94} & \textbf{1080.99} & \textbf{5.99} & \textbf{3.34} & \textbf{0.97} & \textbf{4.68} \\
\hline
\end{tabular}
\end{center}
\end{table*}

Our network is implemented in PyTorch and our network is trained end-to-end. The AdamW optimizer \cite{b19} is adopted with the initial learning rate of $10^{-3}$ and the weight decay is set to $10^{-4}$. Besides, we use the Gradient Centralization \cite{b20} to improve the generalization performance of our network. This operation centralizes the gradient vectors to make them have zero mean.

The error metrics we use in experiments include RMSE ( Root Mean Squared Error [$mm$] ), MAE( Mean Absolute Error [$mm$] ), iRMSE ( Root Mean Squared Error of the Inverse Depth [$\frac{1}{km}$] ), iMAE ( Mean Absolute Error of the Inverse Depth [$\frac{1}{km}$] ), sqErrorRel ( Relative Squared Error [Percent] ) and absErrorRel ( Relative Absolute Error [Percent] ). Since the ground truth in the dataset is semi-dense, we only calculate the loss and error of the valid pixels(depth value $>$ 0) in the ground truth.

\subsection{Comparison on Single-Line Depth Completion}

\begin{figure*}[htb]
  \centering
  \subfigure[RGB Images]{
  \includegraphics[width=0.45\linewidth]{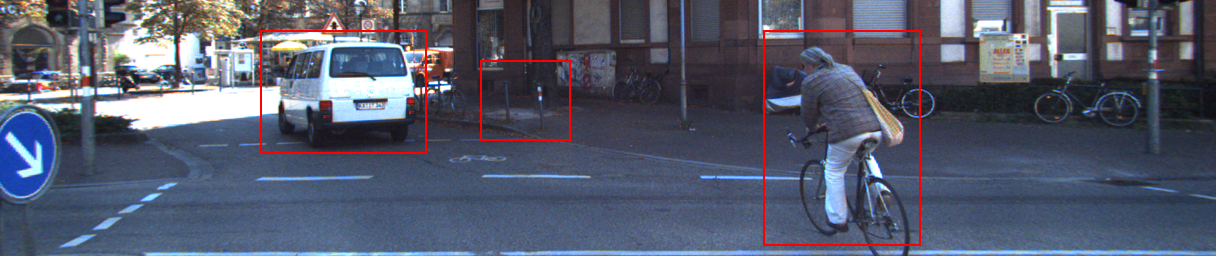}
  \includegraphics[width=0.45\linewidth]{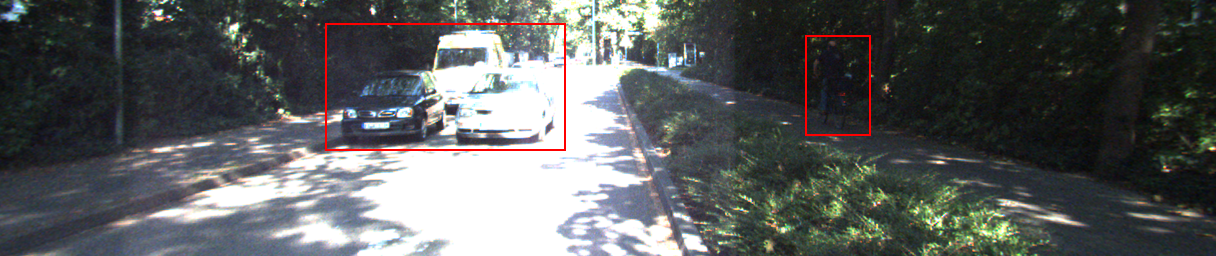}
  }
  \subfigure[Single-Line Depth Maps]{
  \includegraphics[width=0.45\linewidth]{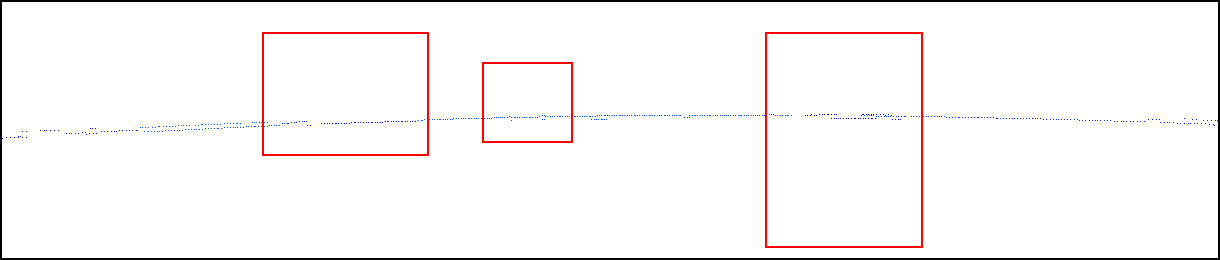}
  \includegraphics[width=0.45\linewidth]{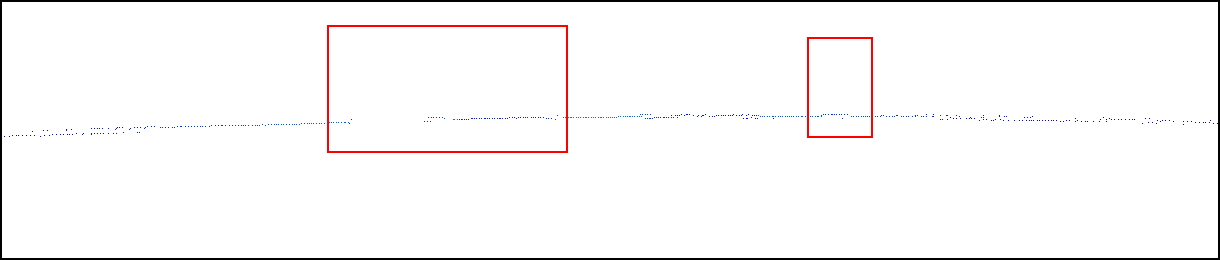}
  }
  \subfigure[Fusion Maps(RGB \& Depth)]{
  \includegraphics[width=0.45\linewidth]{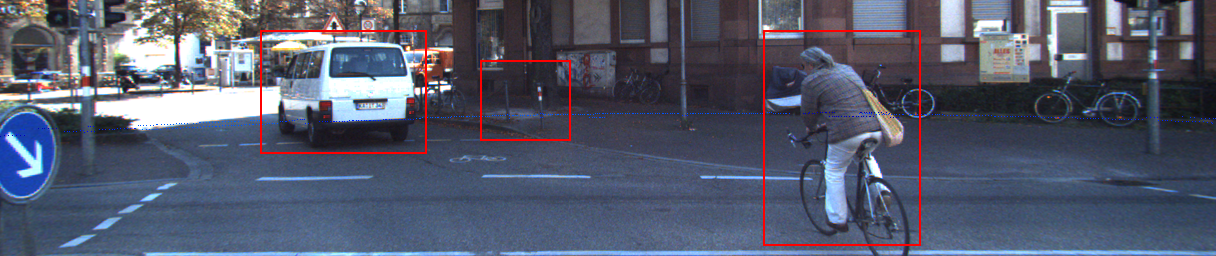}
  \includegraphics[width=0.45\linewidth]{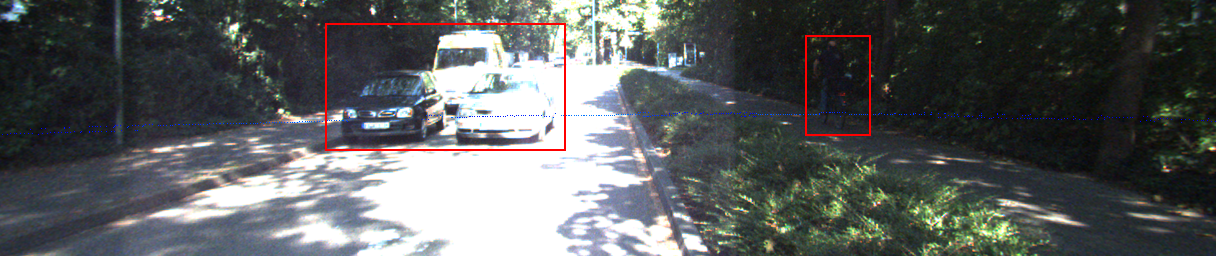}
  }
  \subfigure[BTS(Monocular Depth Estimation)]{
  \includegraphics[width=0.45\linewidth]{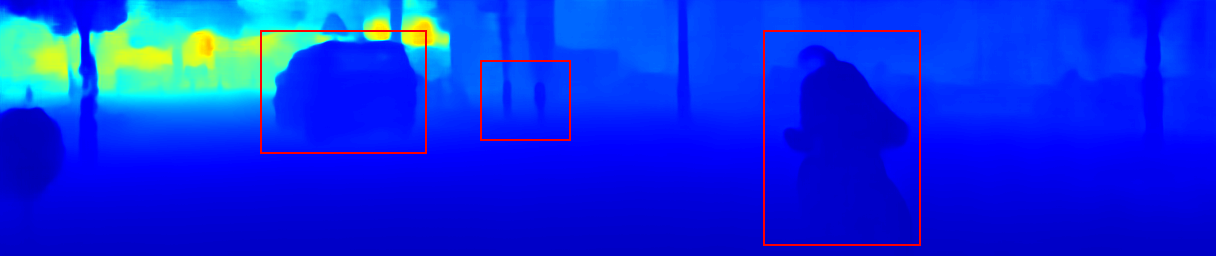}
  \includegraphics[width=0.45\linewidth]{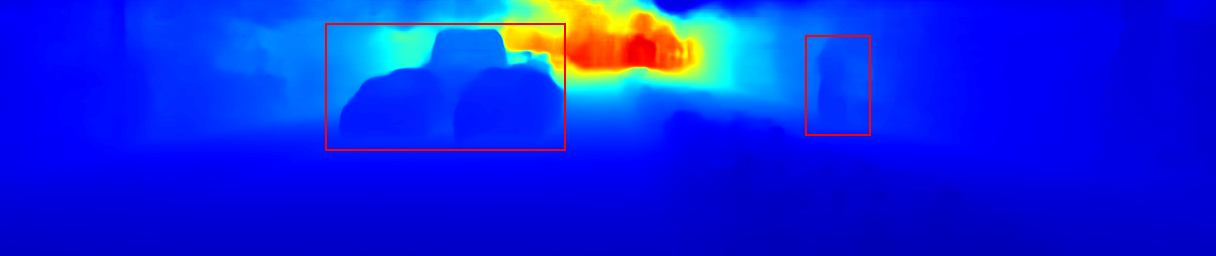}
  }
  \subfigure[Ours(Single-Line Depth Completion)]{
  \includegraphics[width=0.45\linewidth]{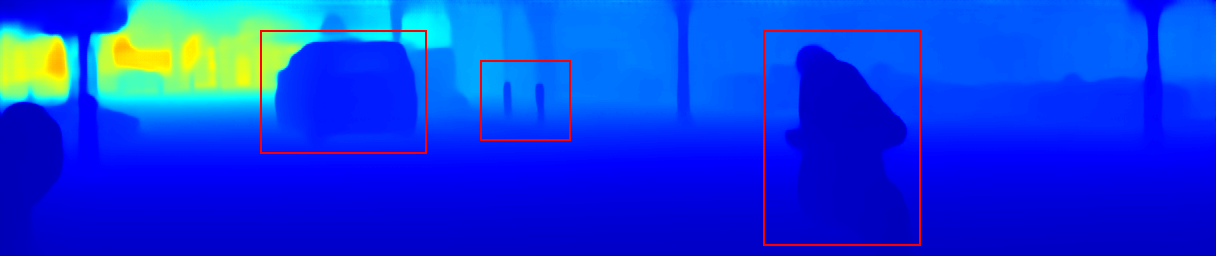}
  \includegraphics[width=0.45\linewidth]{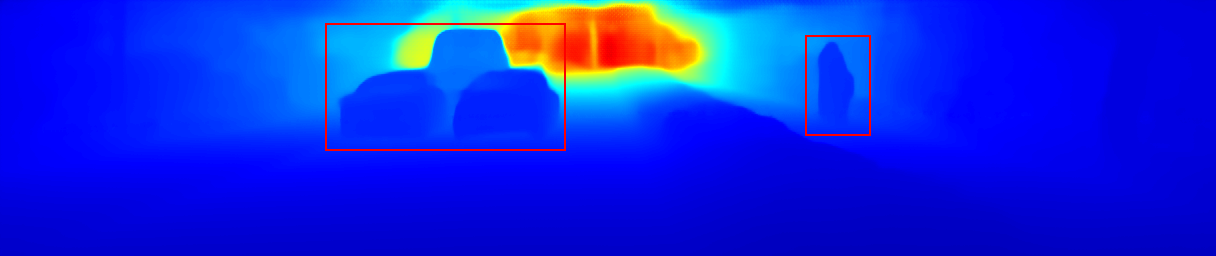}
  }
  \caption{Qualitative comparison with the state-of-the-art monocular depth estimation(BTS). The single-line depth map is only used in the single-line depth completion and not used in monocular depth estimation.}
  \label{fig:qualitative comparison with monocular depth estimation}
\end{figure*}

In this section, we compare our method with other advanced networks designed for depth completion. Although they were previously used for 64-line depth completion, we can also use them in the single-line depth completion task, as long as we replace the 64-line depth map with the single-line depth map during training and inference. To be specific, we retrained and reevaluated these networks using the single-line depth completion dataset we generated to make a fair comparison with us. We use six error metrics that we mentioned above to evaluate networks on the test set on the dataset which we generated.

As shown in Table \ref{single-line depth completion comparison}, our network performs best among all the other networks in each error metric. The error of our network is 8.85$\%$ lower than RGB-guide$\&$certainty \cite{b10} on RMSE. What's more, our network's relative error of 4.68$\%$ also intuitively reflects the accuracy of our network, which means that if the depth value in ground truth is $1m$, the error of our prediction is only $\pm 4.68cm$.

Figure~\ref{fig:qualitative results on single-line depth completion} shows the qualitative comparison of our method to other advanced methods. Our method has the best performance among them. The result on the left shows that the dense depth map generated by our method can more accurately reflect the posture and appearance of the cyclist. The result on the right shows that our method is more accurate in predicting the depth at the boundary of an object. What's more, even in the region which has poor light, our method can still generate the dense depth map accurately.

\subsection{Compare Our Work with Monocular Depth Estimation}

\begin{table*}[htb]\normalsize
\caption{Quantitative comparison with the SOTA monocular depth estimation network(BTS)}
\label{compare with monocular depth estimation}
\begin{center}
\begin{tabular}{|c|l|c|c|c|c|c|c|c|}
\hline
\tabincell{c}{Task} & \tabincell{l}{Network} & \tabincell{c}{PARAM \\ (M) } & \tabincell{c}{RMSE \\ (mm) } & \tabincell{c}{MAE \\ (mm) } & \tabincell{c}{iRMSE \\ (1/km) } & \tabincell{c}{iMAE \\ (1/km) } & \tabincell{c}{sqErrorRel \\ (percent)} & \tabincell{c}{absErrorRel \\ (percent) } \\
\hline
Monocular Depth Estimation & BTS \cite{b14} & 49.53 & 3392.21 & 1525.29 & 8.81 & 5.60 & 1.73 & 7.38 \\
Single-Line Depth Completion & Ours & \textbf{6.39} & \textbf{2921.94} & \textbf{1080.99} & \textbf{5.99} & \textbf{3.34} & \textbf{0.97} & \textbf{4.68} \\
\hline
\end{tabular}
\end{center}
\end{table*}

In addition to the depth completion, there is another low-cost method to acquire dense depth maps, which is called monocular depth estimation. Although this kind of method only needs a single RGB image to realize, the networks of this kind of method often have huge parameter quantity and low precision. In contrast, our method can achieve higher accuracy with fewer parameters by just adding a low-cost single-line LiDAR(\$2,000).

In this section, we will compare our method with the state-of-the-art network(BTS \cite{b14}) in the monocular depth estimation. For fairness, we evaluate our network and theirs on the same dataset. The only difference is that we use the RGB image and single-line depth map as input during training and inference, and they only use a single RGB image as input during training and inference. In the monocular depth estimation, due to the lack of absolute depth values, the output map from the BTS network is a normalized coefficient map. The final depth map used for the evaluation can be obtained by multiplying the coefficient map with the maximum depth value which is $80m$. In the single-line depth completion, absolute depth information has been included in the input, so the direct output is the final depth map.

As shown in Table \ref{compare with monocular depth estimation}, we find that our network far exceeds BTS on all error metrics, and the parameter quantity of our network is also much smaller than BTS. We think it is worthwhile to gain these advantages with the cost of \$2,000.

Figure~\ref{fig:qualitative comparison with monocular depth estimation} shows the qualitative comparison between our method and BTS. According to the dense prediction depth maps, we can find that the depth of the object predicted by BTS is blurry at the boundary, and in the dark area, the object is easily mixed with the background, which is very dangerous in the autonomous driving scene. Our method doesn't have these problems at all.

\subsection{Ablation Study}
To verify the effectiveness of our model's design, we conduct comprehensive ablation experiments.

\subsubsection{Effectiveness of Two Branches}

\begin{figure}[htb]
  \centering
  \subfigure[RGB Images]{
  \includegraphics[width=0.95\linewidth]{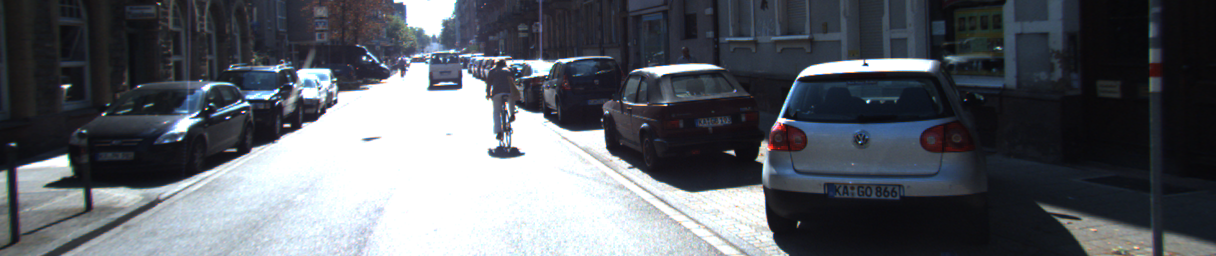}
  }
  \subfigure[Single-Line Depth Maps]{
  \includegraphics[width=0.95\linewidth]{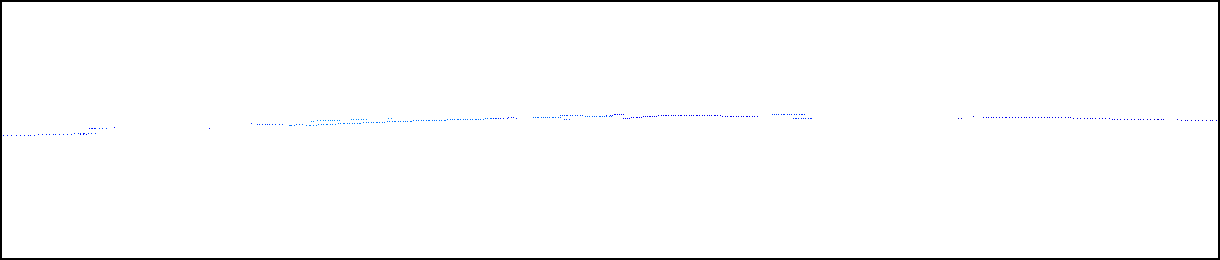}
  }
  \subfigure[Output]{
   \includegraphics[width=0.95\linewidth]{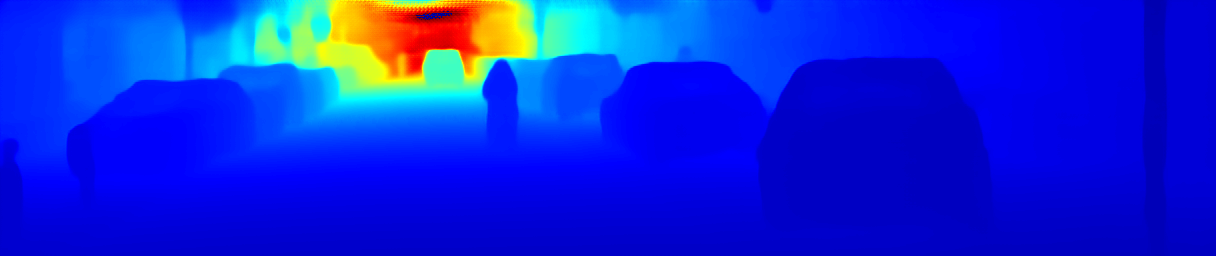}
  }
  \subfigure[Global Confidence]{
  \includegraphics[width=0.95\linewidth]{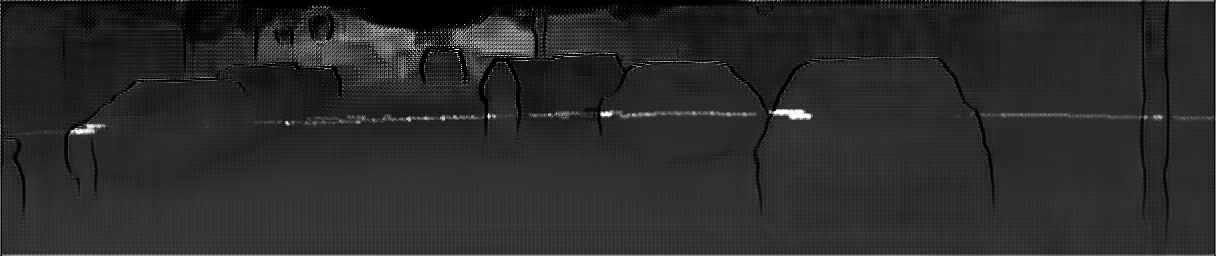}
  }
  \subfigure[Local Confidence]{
  \includegraphics[width=0.95\linewidth]{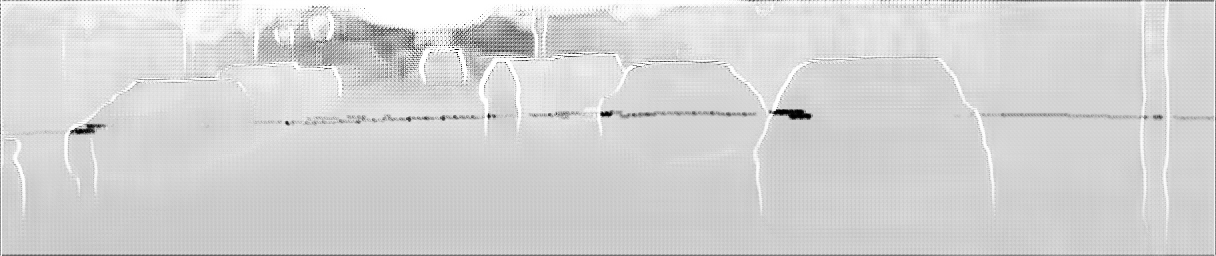}
  }
  \caption{Visualization of global and local confidence maps. Both are grayscale images.}
  \label{fig:confidence ablation visual}
\end{figure}

In our network structure, we used the structure of two branches to achieve single-line depth completion. The ablation experiment of the two branches will be carried out here to demonstrate the roles of the two branches.

Figure~\ref{fig:confidence ablation visual} shows the visualization of global and local confidence maps after the normalization of softmax functions. By observing two confidence maps we can find which branch is dominant in different areas. Both of the two confidence maps are grayscale images, that is, the more white the color in the area, the higher the confidence, and the more black the color in the area, the lower the confidence. After observation, it can be found that the depth values in the valid value region are small. In the area close to the valid values, that is, the area with small depth values, the confidence of the local branch is higher. In the area far away from the valid values, that is, the area with large depth values, the global branch's confidence is higher. Besides, the global branch is more advantageous in the area on valid values. This result is consistent with our previous analysis, which also proves that the depth values of different areas need different branches to complete.

\subsubsection{Effectiveness of SGDUM and VNL}
\begin{table}[htb]
\caption{Ablation study about SGDUM and VNL on the single-line depth completion test set}
\label{Ablation study on SGDUM and VNL}
\begin{center}
\begin{tabular}{|c|c|c|c|c|}
\hline
\tabincell{c}{Model} & \tabincell{c}{RMSE \\ (mm) } & \tabincell{c}{MAE \\ (mm) } & \tabincell{c}{sqErrorRel \\ (percent) } & \tabincell{c}{absErrorRel \\ (percent) } \\
\hline
w/o VNL & 3098.18 & 1208.70 & 1.34 & 5.68 \\
w/o SGDUM & 3013.59 & 1128.84 & 1.01 & 4.82 \\
Ours (Full) & \textbf{2921.94} & \textbf{1080.99} & \textbf{0.97} & \textbf{4.68} \\
\hline
\end{tabular}
\end{center}
\end{table}

In our network, we use SGDUM to make better use of semantic info and add VNL to enforce 3D geometric constraints. To verify if they are effective, we design two ablation experiments in this section. In these two experiments, we train two models that remove SGDUM and VNL respectively, and keep other conditions unchanged. As shown in Tabel \ref{Ablation study on SGDUM and VNL}, when removing SGDUM or VNL, the performance of the model drops significantly than the full model. This demonstrates that the SGDUM and VNL can both improve our model's performance in single-line depth completion.

\subsubsection{Effectiveness of Gradient Centralization}
\begin{table}[htb]
\caption{Ablation study about the Gradient Centralization on the single-line depth completion test set}
\label{Ablation study on Gradient Centralization}
\begin{center}
\begin{tabular}{|c|c|c|c|c|}
\hline
\tabincell{c}{Optimizer} & \tabincell{c}{RMSE \\ (mm) } & \tabincell{c}{MAE \\ (mm) } & \tabincell{c}{sqErrorRel \\ (percent) } & \tabincell{c}{absErrorRel \\ (percent) } \\
\hline
AdamW & 2951.68 & 1094.49 & 1.06 & 4.68 \\
\tabincell{c}{Ours \\ (AdamW-GC)} & \tabincell{c}{\textbf{2921.94}} & \tabincell{c}{\textbf{1080.99}} & \tabincell{c}{\textbf{0.97}} & \tabincell{c}{\textbf{4.68}} \\
\hline
\end{tabular}
\end{center}
\end{table}

We use the AdamW with Gradient Centralization(GC) as our network's optimizer. In this section, we verify its effectiveness in single-line depth completion. We use AdamW and AdamW with Gradient Centralization to train the same model respectively. As shown in Table \ref{Ablation study on Gradient Centralization}, we can find that the performance of the model trained with Gradient Centralization is better on the test set. This demonstrates that this technique can improve the model's generalization performance.

\subsubsection{Study on Multi-Modal Fusion Form}

\begin{table}[htb]
\caption{Ablation study about two multi-modal fusion forms: early fusion and late fusion}
\label{Ablation study on multi-modal fusion form}
\begin{center}
\begin{tabular}{|c|c|c|c|c|}
\hline
\tabincell{c}{Model} & \tabincell{c}{RMSE \\ (mm) } & \tabincell{c}{MAE \\ (mm) } & \tabincell{c}{sqErrorRel \\ (percent) } & \tabincell{c}{absErrorRel \\ (percent) } \\
\hline
Early & 2993.19 & 1114.48 & 1.11 & 4.90 \\
Ours (Late) & \textbf{2921.94} & \textbf{1080.99} & \textbf{0.97} & \textbf{4.68} \\
\hline
\end{tabular}
\end{center}
\end{table}

In this section, we compare two multi-modal data fusion forms: early fusion and late fusion. The operation of early fusion is to extract the feature of RGB-D images with only one encoder in the global branch of our network. In this form, the RGB image and depth map are concatenated directly in the early stage. The operation of late fusion is to use two encoders to extract the feature of RGB images and depth maps respectively and concatenate them in the late stage. As shown in Table \ref{Ablation study on multi-modal fusion form}, the late fusion we use is better. I think it is reasonable because RGB images and depth maps are the data of different modalities. It is inappropriate to use early fusion on them which directly concatenate them together without any processing. The late fusion is more reasonable because we can use different encoders to process them respectively and concatenate their features after processing.

\subsection{Study on 16-Line LiDAR}

\begin{table}[htb]
\caption{Quantitative results on 16-line depth completion}
\label{16-line depth completion}
\begin{center}
\begin{tabular}{|c|c|c|c|c|c|c|c|c|}
\hline
\tabincell{c}{Task} & \tabincell{c}{Cost \\ (\$)} & \tabincell{c}{RMSE \\ (mm) } & \tabincell{c}{MAE \\ (mm) } & \tabincell{c}{sqErrorRel \\ (percent)} & \tabincell{c}{absErrorRel \\ (percent) } \\
\hline
Single-Line & \textbf{2000} & 2921.94 & 1080.99 & 0.97 & 4.68 \\
16-Line & 3500 & \textbf{1209.39} & \textbf{320.06} & \textbf{0.24} & \textbf{1.48} \\
\hline
\end{tabular}
\end{center}
\end{table}

The core goal of us is to obtain accurate and dense depth maps at a low cost. Based on this criterion, we research depth completion based on single-line LiDAR and finally obtain accurate and dense depth maps at a very low cost. However, in some practical application scenarios, sometimes the cost constraint is not so high, at which time we consider using more expensive LiDAR to obtain higher precision. 16-line LiDAR(Ouster) costs about \$3,500, which is much more expensive than single-line LiDAR which costs \$2,000, but it is not unacceptable compared to the \$100,000 cost of 64-line LiDAR. Therefore, we also conducted experiments on depth completion based on 16-line LiDAR.

For 16-line depth completion, the method we use is similar to that of single-line depth completion. The only difference is that in the 16-line depth completion, we process the 64-line depth maps from KITTI to generate the 16-line depth maps. Then we can use the same method to retrain and reevaluate our network. For the fairness of the experiment, we both used our network for 16-line depth completion and single-line depth completion. For the selection of scan lines in the 16-line depth map, the operation we take is to traverse the 64 scan lines in the 64-line depth map and select one every four scan lines.

As shown in Table \ref{16-line depth completion}, the precision of 16-line depth completion is better than that of single-line depth completion in each error metric. This is because, on one hand, compared with the single-line depth map, the 16-line depth map provides richer depth information. On the other hand, the single-line depth map can only provide depth correlation in the horizontal direction, while the 16-line depth map can provide depth correlation both in the horizontal and vertical directions.

Both single-line depth completion and 16-line depth completion are schemes to obtain high-precision depth maps at a low cost. In practical application, the scheme can be selected from them according to the trade-off between cost and precision.

\section{Conclusion}
In this paper, we propose a novel method to generate accurate and dense depth maps from the single-line LiDAR and RGB camera. We generate a single-line depth completion dataset from the KITTI depth completion dataset(64-line) and propose a novel network called SGTBN that achieves state-of-the-art performance in the single-line depth completion. Additionally, we compare our method with the monocular depth estimation. Compared with the monocular depth estimation, our method has significant advantages in precision and model size. At the same time, our method has achieved a relative error of 4.68$\%$ without pre-training, which means that for a depth of $1m$, our error is only $\pm 4.68cm$. Finally, we also studied and analyzed the 16-line depth completion to provide an additional scheme for practical application. With the advantages of high precision and low cost, our method is of great practical value.

\end{document}